\documentclass{article}

\usepackage[final]{neurips_2021}

\usepackage[utf8]{inputenc} 
\usepackage[T1]{fontenc}    
\usepackage{hyperref}       
\usepackage{url}            
\usepackage{booktabs}       
\usepackage{amsfonts}       
\usepackage{nicefrac}       
\usepackage{microtype}      
\usepackage{xcolor}         
\usepackage{rotating}

\title{How not to Lie with a Benchmark: \\Rearranging NLP Leaderboards 
}

\author{Tatiana Shavrina \\
   Sberbank of Russia, SberDevices\\
   Moscow, Russia \\
   AI Research Institute (AIRI) \\
   Moscow, Russia \\
  \texttt{rybolos@gmail.com} \\
   \And
   Valentin Malykh \\
    Huawei Noah's Ark lab \\
   Moscow, Russia \\
   Kazan Federal University \\
   Kazan, Russia \\
   \texttt{valentin.malykh@phystech.edu} \\
}

\begin{document}

\maketitle

\begin{abstract}
Comparison with a human is an essential requirement for a benchmark for it to be a reliable measurement of model capabilities.
Nevertheless, the methods for model comparison could have a fundamental flaw - the arithmetic mean of separate metrics is used for all tasks of different complexity, different size of test and training sets.
 
In this paper, we examine popular NLP benchmarks' overall scoring methods and rearrange the models by geometric and harmonic mean (appropriate for averaging rates) according to their reported results. We analyze several popular benchmarks including GLUE, SuperGLUE, XGLUE, and XTREME. The analysis shows that e.g. human level on SuperGLUE is still not reached, and there is still room for improvement for the current models.

\end{abstract}

\section{Introduction}
The benchmarking approach has a rich history throughout computer science and is now the leading method in machine learning progress validation. In the field of Natural Language Processing (NLP), there exist at least 898 benchmarks\footnote{according to \url{https://paperswithcode.com/area/natural-language-processing}}, the most prominent being GLUE, SuperGLUE, XGLUE, etc., created within a single paradigm. 

The SOTA-centricity of the language model benchmarking has been since criticized as misleading \footnote{\url{https://hackingsemantics.xyz/2019/leaderboards/}}: the battle for percentage fractions leads to extensive improvements -- an increase in the data volume and number of parameters of models, -- rather than an intensive improvement in their architectures. 
The existing solutions suggest Pareto efficiency between the estimated accuracy and computational costs \citep{dodge2019work} and draw attention to the question of model utility versus overall score \citep{ethayarajh2021utility}. The recent work \citep{dehghani2021benchmark} has also stated that many factors, other than fundamental algorithmic superiority, can lead to a method being evaluated as superior, for example, various subsamples of the benchmark tasks' result in different leaderboard arrangements. This forms the conception of the \textit{"benchmark lottery"}, describing the fragility of the main model evaluation instruments.

We state that the delicate question of a general model assessment should not be solved by a simplistic yet incorrect method -- the arithmetic mean over all the results. Stated method does not take into account the scatter of results on various tasks, the differing size of the task test sets (e.g. in SuperGLUE they differ a hundred times, compare 146 test samples in Winograd Schema and 10'000 test samples in ReCoRd\citep {wang2019superglue}), different susceptibility to leaks \citep{elangovan2021memorization}, including the year of creation (Recognizing Textual Entailment data was collected in 2005 \citep{dagan2005pascal}, while BoolQ or CommitmentBank data was collected in 2019\citep {clark2019boolq}, \citep{de2019commitmentbank}).

In this article we present an analysis of the NLP benchmarks' results, using not the arithmetic mean, but the other metrics: geometric mean and harmonic mean. As F1 (harmonic mean) is frequently used to normalize Precision and Recall as they are fractions, optimizing the classifier threshold to maximize F1 leads to a better balance between metrics than the arithmetic mean because it penalizes systems more for the smaller values\citep{sasaki2007truth}. The geometric mean, as noted in the \citep{10.1145/5666.5673}, is the preferred metric to the arithmetic mean in computing performance benchmarks when it comes to normalized values and percentages. 

The results change the usual idea of the models' order on leaderboards: humans still occupy the first place in the intellectual task solving, and the best results (1.5-2\% worse than humans) belong to DeBerta\citep{he2020deberta}, T5+Meena \citep{raffel2020exploring} and McAlbert+DKM models\footnote{\url{https://www.iflytek.com/news/2118}}. Thus, the contribution of this paper is two-fold: 
\begin{enumerate}
    \item we present the reviewed model evaluation technique, core for benchmark design,
    \item we re-arrange the currently existing leaderboards of most popular benchmarks.
\end{enumerate}


\section{Previous Work}
\label{sec:previous_work}
\label{sec:previous_work}
Evaluation and comparison of NLP models beget a rich history, rising with the Turing test \citep{turing2009computing}. The next step in the development and assessment of intelligent systems belongs to the ML-benchmark methodology, which aims to bring the solution of the Natural Language Understanding problem closer - General Language Understanding Evaluation\citep{wang2019glue}.


As stated in~\citep{wang2019superglue}, \textit{"Lacking a fair criterion with which to weight the contributions of each task to the overall score, we opt for the simple approach of weighing each task equally, and for tasks with multiple metrics, first averaging those metrics to get a task score."} All the GLUE-based benchmarks follow this methodology. 

However, apart from the GLUE format, other benchmarks have provided several alternatives to evaluate the overall model contribution. KILT, a Benchmark for Knowledge Intensive Language Tasks~\citep{petroni2020kilt}, 
avoids calculating the overall result, and also do not compare the result with the human level, but only provides metrics for individual tasks. DecaNLP~\citep{mccann2018natural} makes a rating using not the average, but the sum of points for all tasks. This approach allows balancing the contributions of different tasks to the overall metric.

\section{Method}
\label{sec:method}
We arrange the NLP benchmark results using the publicly available model scores for all the tasks to calculate new overall scores.

Other available options from Pythagorean means - harmonic mean and geometric mean - can also be considered: we have centered our research around 2 simple statistics that are widely used for averaging fractions~\citep{stats1969} or normalized values~\citep{10.1145/5666.5673} among the possible alternatives. 
We did not consider other measures of central tendency, like median and mode, as the averaged samples more often consist of about 10 measurements, and on them such metrics can give the same results for competing systems.
\begin{itemize}
    \item The arithmetic mean (AM) desribed in eq.~\ref{eqn:am} is calculated as the sum of the task scores (Xs) divided by the total number of tasks, referred to as N.
    \item The geometric mean (GM) desribed in eq.~\ref{eqn:gm} is calculated as the N-th root of the product of all task scores (with the above conditions), where N is the number of values.
    \item The harmonic mean (HM) desribed in eq.~\ref{eqn:hm} is calculated as the number of values N divided by the sum of the reciprocal of the values
\end{itemize}

\begin{equation}
{\displaystyle AM={\frac {1}{n}}\sum _{i=1}^{n}x_{i}={\frac {x_{1}+x_{2}+\cdots +x_{n}}{n}}}
\label{eqn:am}
\end{equation}
\begin{equation}
{\displaystyle GM= \left(\prod _{i=1}^{n}x_{i}\right)^{\frac {1}{n}}={\sqrt[{n}]{x_{1}x_{2}\cdots x_{n}}}}
\label{eqn:gm}
\end{equation}

\begin{equation}
{\displaystyle HM=\frac {N}{{\frac {1}{x_{1}}}+{\frac {1}{x_{2}}}+\cdots +{\frac {1}{x_{N}}}}}
\label{eqn:hm}
\end{equation}

The sections below present the results of the leaderboard re-weighting as of May 2021. 

\subsection{Reevaluating the Benchmarks}
\label{reeval}

The GLUE \textit{(11 tasks for English)}, SuperGLUE \textit{(10 tasks for English)}, XGLUE \textit{(11 tasks for 19 languages)}, and XTREME \textit{(4 tasks for 40 languages)} provide different scoring metrics for each task, including Accuracy, F1, Matthew’s correlation coefficient, Exact Match, while the overall score is calculated by their simple average. In cases like these, the geometric mean is appropriate when the data contains values with different units of measure~\citep{stats1969}.

The harmonic mean of the task results as a better overall metric has the same grounding as introduction of the F-1 measure over precision and recall \citep{sasaki2007truth}: the harmonic mean is more intuitive than the arithmetic mean when computing a mean of ratios. Given the set of metrics with a large scatter, the harmonic mean will be less than the arithmetic mean, penalizing the system more for the errors made.


The harmonic mean is the appropriate mean if the data is comprised of rates, while the geometric mean is used as an unbiased estimation when working with normalized ratios, for example, in finance~\citep{dittmann2008biases} or computing benchmarks~\citep{10.1145/5666.5673}. 

However, their applicability to a better summarization of the model performance to a single number has been widely discussed, see \citep{smith1988characterizing}, discussing performance computing: 
\begin{itemize}
    \item the \textbf{harmonic mean} is considered the appropriate metric to summarize benchmark results expressed as rates,
    \item while \textbf{geometric mean} is applicable in case of the use of performance numbers that are normalized with respect to one of the results being compared (see \ref{sec:results}),
    \item and \textbf{arithmetic mean} should not be used as a summarizing metric with rates, making it the worst choice for results accumulation.
\end{itemize}

\subsection{GLUE}
GLUE benchmark~\citep{wang2019glue} combines 11 tasks in various text classification and question answering.

\textbf{Overall score:} average of all the task results. If task has 2 main metrics, these metrics are averaged, then added to the overall average.

\textbf{Human evaluation:} collected on reported human performance numbers from original datasets, not exceeding 200 examples (heavily criticised in \citep{nangia2019human}). The human baseline performance on the diagnostic set was provided by the project authors with the help of six NLP researchers annotating 50 randomly selected sentence pairs. 

\textbf{Rearranging the scores:} the results of geometric and harmonic mean rearrangement are presented in Tab.~\ref{table:glue}. GLUE benchmark seem to be the most reordered of all the ratings considered: the best result by geometric and harmonic means belongs to humans, DeBerta and McAlbert+DKM got a 1 point demotion, and the other models got severely rearranged  their places. 

\begin{table*}[htp!]
\centering
\tiny
\begin{tabular}{lllllllllllll}
\hline
\textbf{N} & \textbf{Name}                                               & \textbf{AM} & \textbf{HM} & \textbf{GM} & \textbf{CoLA} & \textbf{SST-2} & \textbf{\begin{tabular}[c]{@{}l@{}}MRPC\\ Mean\end{tabular}} & \textbf{\begin{tabular}[c]{@{}l@{}}STS-B\\ Mean\end{tabular}} & \textbf{\begin{tabular}[c]{@{}l@{}}QQP\\ Mean\end{tabular}} & \textbf{\begin{tabular}[c]{@{}l@{}}MNLI\\ m\end{tabular}} & \textbf{\begin{tabular}[c]{@{}l@{}}MNLI\\ mm\end{tabular}} & \textbf{QNLI} \\ \hline
16            & Human                                                       & 87.10                                                         & 86.16                                                        & 86.91                                                        & 66.40         & 97.80          & 83.55                                                        & 92.65                                                         & 69.95                                                       & 92.00                                                     & 92.80                                                      & 91.20         \\
1             & DeBERTa                                                     & 90.80                                                         & 84.78                                                        & 86.25                                                        & 71.50         & 97.50          & 93.00                                                        & 92.75                                                         & 83.50                                                       & 91.90                                                     & 91.60                                                      & 99,20         \\
2             & \begin{tabular}[c]{@{}l@{}}Mac\\ Albert\\ +DKM\end{tabular} & 90.70                                                         & 84.70                                                        & 86.13                                                        & 74.80         & 97.00          & 93.55                                                        & 92.70                                                         & 82,65                                                       & 91.30                                                     & 91.10                                                      & 97.80         \\
6             & T5                                                          & 90.30                                                         & 84.48                                                        & 85.92                                                        & 71.60         & 97.50          & 91.60                                                        & 92.95                                                         & 82.85                                                       & 92.20                                                     & 91.90                                                      & 96.90         \\
4             & PING-AN                                                     & 90.60                                                         & 84.26                                                        & 85.83                                                        & 73.50         & 97.20          & 93.00                                                        & 92.70                                                         & 83.55                                                       & 91.60                                                     & 91.30                                                      & 97.50         \\
5             & ERNIE                                                       & 90,40                                                         & 84.27                                                        & 85.75                                                        & 74,40         & 97.50          & 92.45                                                        & 92.80                                                         & 83.05                                                       & 91.40                                                     & 91.00                                                      & 96.60        
\end{tabular}
\caption{Top results of ranking GLUE benchmark with geometric mean. N -- original model rank on the leaderboard. 
MNLI m and MNLI mm correspond to MultiNLI Matched \& MultiNLI Mismatched, other task abbreviations correspond to their GLUE leaderboard designations accordingly.}
\label{table:glue}
\end{table*}

\subsection{SuperGLUE}
SuperGLUE~\citep{wang2019superglue} is the sophisticated version of the GLUE benchmark, combining 10 tasks with a higher demand for higher intellectual abilities. Task data must is available under various licenses that allow use and redistribution for research purposes.

\textbf{Overall score:} average of all the task results. If task has 2 main metrics, these metrics are averaged, then added to the overall average.

\textbf{Human evaluation:} ready-made estimates for WiC, MultiRC, RTE, and ReCoRD datasets, the other tasks being evaluated by the project creators with the help of crowdworker annotators through Amazon’s Mechanical Turk.

\textbf{Rearranging the scores:} Re-weighting the results using the geometric mean and harmonic mean again makes significant changes to the original ranking: the top-3 result (human) is ranked top-1, the DeBerta and T5 models are shifted down 1 position, PAI ALbert and Nezha Plus models swap their places, see Tab.~\ref{table:superglue}.

\begin{table*}[htp!]
\centering
\tiny
\begin{tabular}{lllllllllllllll}
\hline
\textbf{N} & \textbf{Model}                                       & \textbf{AM} & \textbf{HM} & \textbf{GM} & \textbf{BoolQ} & \textbf{\begin{tabular}[c]{@{}l@{}}CB\\ Mean\end{tabular}} & \textbf{COPA} & \textbf{\begin{tabular}[c]{@{}l@{}}MRC\end{tabular}} & \textbf{\begin{tabular}[c]{@{}l@{}}RCD\end{tabular}} & \textbf{RTE} & \textbf{WiC} & \textbf{WSC} & \textbf{AX-b} & \textbf{\begin{tabular}[c]{@{}l@{}}AX-g\\ mean\end{tabular}} \\ \hline
3             & Human                                                & 89,80                                                         & 87,96                                                        & 88,73                                                        & 89,00          & 97,35                                                      & 100,00        & 66,85                                                           & 91,50                                                          & 93,60        & 80,00        & 100,00       & 76,6          & 99,5                                                         \\
1             & DeBERTa                                              & 90,30                                                         & 86,89                                                        & 87,60                                                        & 90,40          & 96,65                                                      & 98,40         & 75,95                                                           & 94,30                                                          & 93,20        & 77,50        & 95,90        & 66,7          & 93,55                                                        \\
2             & \begin{tabular}[c]{@{}l@{}}T5+\\ Meena\end{tabular}  & 90,20                                                         & 86,42                                                        & 87,10                                                        & 91,30          & 96,70                                                      & 97,40         & 75,65                                                           & 93,85                                                          & 92,70        & 77,90        & 95,90        & 66,5          & 89,35                                                        \\
4             & T5                                                   & 89,30                                                         & 85,89                                                        & 86,57                                                        & 91,20          & 95,35                                                      & 94,80         & 75,70                                                           & 93,75                                                          & 92,50        & 76,90        & 93,80        & 65,6          & 92,3                                                         \\
6             & \begin{tabular}[c]{@{}l@{}}PAI\\ Albert\end{tabular} & 86,10                                                         & 85,24                                                        & 85,78                                                        & 88,10          & 94,40                                                      & 91,80         & 69,65                                                           & 88,65                                                          & 88,80        & 74,10        & 93,20        & 75,6          & 98,75                                       
 \\
5             & \begin{tabular}[c]{@{}l@{}}Nezha\\ plus\end{tabular} & 86,70                                                                                                                                                           & 81,30                                                                                                                                                          & 82,29                                                                                                                                                          & 87,80                                    & 95,20                                                                                                                                                        & 93,60                                   & 69,85                                                                                                                      & 89,85                                                                                                             & 89,10                         & 74,60                         & 93,20                         & 58,00                          & 80,75                                       
\end{tabular}
\caption{Top results of ranking SuperGLUE benchmark with geometric mean. N -- original model rank on the leaderboard MRC stands for MultiRC averaged metric, RCD - ReCoRD averaged metric.}
\label{table:superglue}
\end{table*}

\subsection{XTREME}
The XTREME benchmark~\citep{hu2020xtreme}
 covers 40 typologically diverse languages from 12 language families and includes 9 tasks that require analysis of different levels of syntax or semantics.

\textbf{Overall score:} 2-step averaging: 1) calculating average for each task on all languages 2) calculating average on all tasks.

\textbf{Human evaluation:} 2-step averaging:
\begin{enumerate}
    \item step 1: ready-made estimates from the original datasets taken and extrapolated to all unestimated languages; besides, for some datasets there were no original estimates provided (POS) and an empirical estimate of 97\% was taken based on \citep{manning2011part}; no estimates for NER and sentence retrieval tasks;
    \item step 2: all the task results averaged together.
\end{enumerate}

\textbf{Rearranging the scores:} the results of applying the geometric mean and harmonic mean did not change the current ranking of the models - the quality spread between them is high enough for the metrics averaging them to retain the current order.

\begin{table*}[htp!]
\centering
\small
\begin{tabular}{lllllllll}
\hline
\textbf{N} & \textbf{Model} & \textbf{AM} & \textbf{HM} & \textbf{GM} & \textbf{\begin{tabular}[c]{@{}l@{}}Sentence-pair\\ Classification\end{tabular}} & \textbf{\begin{tabular}[c]{@{}l@{}}Structured\\ Prediction\end{tabular}} & \textbf{\begin{tabular}[c]{@{}l@{}}Question\\ Answering\end{tabular}} & \textbf{\begin{tabular}[c]{@{}l@{}}Sentence\\ Retrieval\end{tabular}} \\ \hline
1          & Human          & 93,30                                                         & 93,13                                                        & 93,21                                                        & 95,10                                                                           & 97,00                                                                    & 87,80                                                                 & 0.00001                                                               \\
2          & VECO           & 81,10                                                         & 81,27                                                        & 81,70                                                        & 88,60                                                                           & 75,40                                                                    & 72,40                                                                 & 92,10                                                                 \\
3          & ERNIE-M        & 80,90                                                         & 81,11                                                        & 81,52                                                        & 87,90                                                                           & 75,60                                                                    & 72,30                                                                 & 91,90                                                                 \\
4          & T-ULRv2        & 80,70                                                         & 80,91                                                        & 81,25                                                        & 88,80                                                                           & 75,40                                                                    & 72,90                                                                 & 89,30                                                                 \\
5          & Anonymous3     & 79,90                                                         & 80,12                                                        & 80,50                                                        & 88,20                                                                           & 74,60                                                                    & 71,70                                                                 & 89,00                                                                 \\
6          & Polyglot       & 77,80                                                         & 78,02                                                        & 78,56                                                        & 87,80                                                                           & 72,90                                                                    & 67,40                                                                 & 88,30                                                                
\end{tabular}
\caption{Top results of ranking XTREME benchmark with geometric mean. N -- original model rank on the leaderboard; the averaged task scores are shown by the column markings.}
\label{table:xtreme}
\end{table*}

\subsection{XGLUE}
The XGLUE benchmark\citep{liang2020xglue}
consists of 11 problems in 19 languages and evaluates the performance of multilingual pre-trained systems in terms of their ability to cross-language understanding and natural language generation. 

\textbf{Overall score:} 2-step averaging: 1) calculating average for each task on all languages 2) calculating average on all tasks.

\textbf{Human evaluation:} not provided. 

\textbf{Rearranging the scores:} Since the human level is not measured in the benchmark, we can only compare the 2 present models with each other. Tab.~\ref{table:xglue} shows the results - the difference in the quality of the models is large enough to preserve their ranking on all averaging metrics.

\begin{table*}[htp!]
\centering 
\tiny
\begin{tabular}{llllllllllllll}
\hline
\textbf{N} & \textbf{Model}                                              & \textbf{AM} & \textbf{HM} & \textbf{GM}  & \textbf{NER} & \textbf{POS} & \textbf{NC} & \textbf{MLQA} & \textbf{XNLI} & \textbf{PAWS-X} & \textbf{QADSM} & \textbf{WPR} & \textbf{QAM} \\ \hline
1          & FILTER                                                      & 80.10                                                         & 79.61                                                         & 79.86                                                         & 82.60        & 81.60        & 83.50       & 76.20         & 83.90         & 93.80           & 71.40          & 74.70        & 73.40        \\
2          & \begin{tabular}[c]{@{}l@{}}Unicoder\\ Baseline\end{tabular} & 76.10                                                         & 75.45                                                         & 75.80                                                         & 79.70        & 79.60        & 83.50       & 66.00         & 75.30         & 90.10           & 68.40          & 73.90        & 68.90        \\ \cline{1-7}
\textbf{N} & \textbf{Model}                                              & \textbf{AM} & \textbf{HM} & \textbf{GM} & \textbf{QG}  & \textbf{NTG} &             &               &               &                 &                &              &              \\ \cline{1-7}
1          & \begin{tabular}[c]{@{}l@{}}Unicoder\\ Baseline\end{tabular} & 10.70                                                         & 10.65                                                         & 9.10                                                           & 10.60        & 10.70        &             &               &               &                 &                &              &              \\
2          & MP-Tune                                                     & 8.70                                                          & 8.70                                                          & 7.10                                                           & 8.10         & 9.40         &             &               &               &                 &                &              &             
\end{tabular}
\caption{Top results of ranking XGLUE benchmark with geometric mean. N -- original model rank on the leaderboard; the first 2 rows correspond to NLU tasks, the last 2 rows - to the NLG tasks.} 
\label{table:xglue}
\end{table*}

\section{Results and Discussion}
\label{sec:results}

The results show that the ranking of results within a single leaderboard can fluctuate significantly. So, in GLUE, the first place in terms of the harmonic and geometric mean belongs to the result occupying the 16th line in the arithmetic mean. In SuperGLUE, the permutation is not so striking - the third result is on the 1st place. On the XTREME and XGLUE benchmarks system ranking is preserved.

Since all three averaging metrics considered are subject to different biases, we present the statistical measurements of the top-3 SuperGLUE results in Tab.~\ref{table:stats}. Human results have the highest total points for all tasks (as in the DecaNLP methodology), while the standard deviation and variance are greater than top-2 and top-3 models. 

\begin{table*}[htp!]
\centering

\begin{tabular}{llllrrr}
\hline
\textbf{Model} & \textbf{AM} & \textbf{GM} & \textbf{HM} & \multicolumn{1}{l}{\textbf{Sum}} & \multicolumn{1}{l}{\textbf{Var}} & \multicolumn{1}{l}{\textbf{Std}} \\ \hline
Human          & 89,8                                                          & \textbf{88,73}                                               & \textbf{87,96}                                               & \textbf{894,40}                  & 130,31                           & 11,42                            \\
DeBerta           & \textbf{90,3}                                                 & 87,60                                                        & 86,89                                                        & 882,55                           & 117,46                           & 10,84                            \\
T5 + Meena        & 90,2                                                          & 87,10                                                        & 86,42                                                        & 877,25                           & \textbf{112,39}                  & \textbf{10,60}                  
\end{tabular}
\caption{Measuring the statistics of the top-3 SuperGLUE results. Sum is a sum of all the task scores, Var and Std are variance and standard deviation on the task scores respectively. Notable results are highlighted in bold. }
\label{table:stats}
\end{table*}

The following topics remain debatable and need special attention of the community:
\begin{enumerate}
    \item Different metrics for obtaining the average value (arithmetic, geometric, harmonic) have different restrictions on the accepted values (for example, not every one can take negative or zero values). At the same time, metrics that take zero and negative values are actively used in measuring various skills - MCC metric on SuperGLUE diagnostics can be negative, other metrics can be equal to or greater than zero, and they are averaged altogether.
    \item Correct averaging of the overall score for multilingual benchmarks creates additional problems while performing the averaging operation in 2 stages: for all languages and all tasks. 
    \item In addition to the problem of the main averaging metric, we left outside of the scope the problem that was also discovered within the framework of this study: human benchmark scores on various tasks were obtained in a very different way.

\end{enumerate}

\section{Conclusion}
\label{sec:conclusion}
\label{sec:conclusion}
In this paper we present an alternative method to arrange the popular NLP benchmark results, elaborating on several task evaluation. 
We analyze popular benchmark averaging methods and provide new insight into model comparison. Namely, we obtain the following results: 
\begin{itemize}
    \item for popular benchmarks GLUE and SuperGLUE we can conclude that their overall score is subject to bias due to outliers; the alternative arrengement methods end with significantly different ordering of the results;
    \item rebuilding leaderboards using other metrics (geometric or harmonic mean) allows one to conclude that human result is the first in the rankings;
    \item in XGLUE leaderboard human result is obtained by extrapolation from one language to others, while in practice the level of problem-solving by native speakers of different languages varies;
    \item the last finding could be extended to other multilingual benchmarks also.
\end{itemize}
We believe that an unbiased generalization of the benchmark scores is a necessity for the community to target language modelling. The tracking of the complex process of model improvement over time can require the improvement of the benchmark design itself.

The expansion towards multilingualism and multimodality of the new benchmark practices plus new models makes the stated problem more urgent and we hope our help to foster research in this direction.


\bibliography{refer}
\bibliographystyle{acl_natbib}



\_ \\\\\

\section*{Appendix}

See tables below 

\subsection{Appendix 1}
\begin{sidewaystable}
\centering
\caption{SuperGLUE benchmark results normalized to human level: geometric mean is the main metric}\label{tab:norm}
\begin{tabular}{llllllllllllllll}
\textbf{Rank} & \textbf{\begin{tabular}[c]{@{}l@{}}SuperGLUE\\ Model\end{tabular}} & \textbf{\begin{tabular}[c]{@{}l@{}}HM\end{tabular}} & \textbf{\begin{tabular}[c]{@{}l@{}}GM\end{tabular}} & \textbf{\begin{tabular}[c]{@{}l@{}}AM\end{tabular}} & \textbf{BoolQ} & \textbf{\begin{tabular}[c]{@{}l@{}}CB\\ mean\end{tabular}} & \textbf{COPA}                                                & \textbf{\begin{tabular}[c]{@{}l@{}}MultiRC\\ Mean\end{tabular}} & \textbf{\begin{tabular}[c]{@{}l@{}}ReCoRD\\ mean\end{tabular}} & \textbf{RTE}    & \textbf{WiC}     & \textbf{WSC}  & \textbf{AX-b} & \textbf{\begin{tabular}[c]{@{}l@{}}AX-g\\ mean\end{tabular}} &             \\ \cline{1-15}
3             & Human                                                              & \textbf{87,9 (1)}                                                   & \textbf{88,7 (1)}                                                   & \textbf{89,8  (1)}                                                    & 89 (1)         & 97,35  (1)                                                 & 100  (1)                                                     & 66,85 (1)                                                       & 91,5 (1)                                                       & 93,6 (1)        & 80 (1)           & 100 (1)       & 76,6  (1)     & 99,5 (1)                                                     &             \\
1             & DeBERTa                                                            & \textbf{0,985}                                               & \textbf{0,987}                                               & \textbf{0,989}                                                & 1,016          & 0,993                                                      & 0,984                                                        & 1,136                                                           & 1,031                                                          & 0,996           & 0,969            & 0,959         & 0,871         & 0,940                                                        &             \\
2             & T5 + Meena                                                         & \textbf{0,979}                                               & \textbf{0,982}                                               & \textbf{0,984}                                                & 1,026          & 0,993                                                      & 0,974                                                        & 1,132                                                           & 1,026                                                          & 0,990           & 0,974            & 0,959         & 0,868         & 0,898                                                        &             \\
4             & T5                                                                 & \textbf{0,973}                                               & \textbf{0,976}                                               & \textbf{0,978}                                                & 1,025          & 0,979                                                      & 0,948                                                        & 1,132                                                           & 1,025                                                          & 0,988           & 0,961            & 0,938         & 0,856         & 0,928                                                        &             \\ \hline
\textbf{Rank} & \textbf{\begin{tabular}[c]{@{}l@{}}GLUE\\ Model\end{tabular}}      & \textbf{\begin{tabular}[c]{@{}l@{}}HM\end{tabular}} & \textbf{\begin{tabular}[c]{@{}l@{}}GM\end{tabular}} & \textbf{\begin{tabular}[c]{@{}l@{}}AM\end{tabular}} & \textbf{CoLA}  & \textbf{SST-2}                                             & \textbf{\begin{tabular}[c]{@{}l@{}}MRPC\\ Mean\end{tabular}} & \textbf{\begin{tabular}[c]{@{}l@{}}STS-B\\ Mean\end{tabular}}   & \textbf{\begin{tabular}[c]{@{}l@{}}QQP\\ Mean\end{tabular}}    & \textbf{MNLI-m} & \textbf{MNLI-mm} & \textbf{QNLI} & \textbf{RTE}  & \textbf{WNLI}                                                & \textbf{AX} \\ \hline
16            & Human                                                              & \textbf{86,1 (1)}                                                   & \textbf{86,9 (1)}                                                   & \textbf{87,1 (1)}                                                    & 66,4 (1)       & 97,8 (1)                                                   & 83,55 (1)                                                    & 92,65 (1)                                                       & 69,95 (1)                                                      & 92 (1)          & 92,8 (1)         & 91,2 (1)      & 93,6 (1)      & 95,9 (1)                                                     & 76,6 (1)    \\
1             & DeBERTa                                                            & \textbf{0,995}                                               & \textbf{1,004}                                               & \textbf{1,012}                                                & 1,077          & 0,997                                                      & 1,113                                                        & 1,001                                                           & 1,194                                                          & 0,999           & 0,987            & 1,088         & 0,996         & 0,985                                                        & 0,695       \\
2             & McAlbert                                                        & \textbf{0,993}                                               & \textbf{1,002}                                               & \textbf{1,011}                                                & 1,127          & 0,992                                                      & 1,120                                                        & 1,001                                                           & 1,182                                                          & 0,992           & 0,982            & 1,072         & 0,983         & 0,985                                                        & 0,687       \\
6             & T5                                                                 & \textbf{0,991}                                               & \textbf{1,000}                                               & \textbf{1,008}                                                & 1,078          & 0,997                                                      & 1,096                                                        & 1,003                                                           & 1,184                                                          & 1,002           & 0,990            & 1,063         & 0,991         & 0,985                                                        & 0,693      
\end{tabular}
\end{sidewaystable}

\subsection{Appendix 2}
\begin{sidewaystable}
\centering
\caption{GLUE results with all metrics for the tasks, including tasks with double metrics. }
\label{table:app2}
\begin{tabular}{|l|l|l|l|l|l|lll|lll|lll|l|l|l|l|l|l|}
\hline
N & Model                                                 & \textbf{\begin{tabular}[c]{@{}l@{}}GM\end{tabular}} & \textbf{\begin{tabular}[c]{@{}l@{}}AM\end{tabular}} & CoLA & SST-2 & \begin{tabular}[c]{@{}l@{}}MRPC\\ F1\end{tabular} & \begin{tabular}[c]{@{}l@{}}MRPC\\ Acc\end{tabular} & \begin{tabular}[c]{@{}l@{}}MRPC\\ Mean\end{tabular} & \begin{tabular}[c]{@{}l@{}}STS-B\\ Prs\end{tabular} & \begin{tabular}[c]{@{}l@{}}STS-B\\ Spr\end{tabular} & \begin{tabular}[c]{@{}l@{}}STS-B\\ Mean\end{tabular} & \begin{tabular}[c]{@{}l@{}}QQP\\ F1\end{tabular} & \begin{tabular}[c]{@{}l@{}}QQP\\ Acc\end{tabular} & \begin{tabular}[c]{@{}l@{}}QQP\\ Mean\end{tabular} & \begin{tabular}[c]{@{}l@{}}MNLI\\ m\end{tabular} & \begin{tabular}[c]{@{}l@{}}MNLI\\ mm\end{tabular} & QNLI & RTE  & WNLI & AX   \\ \hline
16   & Human                                                 & \textbf{86,906}                                                & 87,1                                                         & 66,4 & 97,8  & 86,3                                              & 80,8                                               & 83,55                                               & 92,7                                                & 92,6                                                & 92,65                                                & 59,5                                             & 80,4                                              & 69,95                                              & 92                                               & 92,8                                              & 91,2 & 93,6 & 95,9 & -    \\ \cline{1-4}
1    & DeBERTa                                               & \textbf{86,248}                                                & 90,8                                                         & 71,5 & 97,5  & 94                                                & 92                                                 & 93                                                  & 92,9                                                & 92,6                                                & 92,75                                                & 76,2                                             & 90,8                                              & 83,5                                               & 91,9                                             & 91,6                                              & 99,2 & 93,2 & 94,5 & 53,2 \\ \cline{1-4}
2    & \begin{tabular}[c]{@{}l@{}}HFL\\ iFLYTEK\end{tabular} & \textbf{86,127}                                                & 90,7                                                         & 74,8 & 97    & 94,5                                              & 92,6                                               & 93,55                                               & 92,8                                                & 92,6                                                & 92,7                                                 & 74,7                                             & 90,6                                              & 82,65                                              & 91,3                                             & 91,1                                              & 97,8 & 92   & 94,5 & 52,6 \\ \cline{1-4}
6    & T5                                                    & \textbf{85,915}                                                & 90,3                                                         & 71,6 & 97,5  & 92,8                                              & 90,4                                               & 91,6                                                & 93,1                                                & 92,8                                                & 92,95                                                & 75,1                                             & 90,6                                              & 82,85                                              & 92,2                                             & 91,9                                              & 96,9 & 92,8 & 94,5 & 53,1 \\ \cline{1-4}
4    & PING-AN                                               & \textbf{85,827}                                                & 90,6                                                         & 73,5 & 97,2  & 94                                                & 92                                                 & 93                                                  & 93                                                  & 92,4                                                & 92,7                                                 & 76,1                                             & 91                                                & 83,55                                              & 91,6                                             & 91,3                                              & 97,5 & 91,7 & 94,5 & 51,2 \\ \cline{1-4}
5    & ERNIE                                                 & \textbf{85,754}                                                & 90,4                                                         & 74,4 & 97,5  & 93,5                                              & 91,4                                               & 92,45                                               & 93                                                  & 92,6                                                & 92,8                                                 & 75,2                                             & 90,9                                              & 83,05                                              & 91,4                                             & 91                                                & 96,6 & 90,9 & 94,5 & 51,7 \\ \hline
\end{tabular}
\end{sidewaystable}

\subsection{Appendix 3}
\begin{sidewaystable}
\centering
\caption{SuperGLUE results with all metrics for the tasks, including tasks with double metrics. }
\label{table:app3}
\begin{tabular}{|l|l|l|l|l|lll|l|lll|lll|l|l|l|}
\hline
\textbf{Rank} & \textbf{Model}                                       & \textbf{\begin{tabular}[c]{@{}l@{}}GM\end{tabular}} & \textbf{\begin{tabular}[c]{@{}l@{}}AM\end{tabular}} & \textbf{BoolQ} & \begin{tabular}[c]{@{}l@{}}CB\\ F1\end{tabular} & \begin{tabular}[c]{@{}l@{}}CB\\ Acc\end{tabular} & \textbf{\begin{tabular}[c]{@{}l@{}}CB\\ mean\end{tabular}} & \textbf{COPA} & \begin{tabular}[c]{@{}l@{}}Multi\\ RC F1\end{tabular} & \begin{tabular}[c]{@{}l@{}}Multi\\ RC Acc\end{tabular} & \textbf{\begin{tabular}[c]{@{}l@{}}Multi\\ RC Mean\end{tabular}} & \begin{tabular}[c]{@{}l@{}}ReCoRD\\ F1\end{tabular} & \begin{tabular}[c]{@{}l@{}}ReCoRD\\ Acc\end{tabular} & \textbf{\begin{tabular}[c]{@{}l@{}}ReCoRD\\ mean\end{tabular}} & \textbf{RTE} & \textbf{WiC} & \textbf{WSC} \\ \hline
3             & Human                                                & \textbf{88,729}                                               & 89,8                                                         & 89             & 95,8                                            & 98,9                                             & 97,35                                                      & 100           & 81,8                                                  & 51,9                                                   & 66,85                                                            & 91,7                                                & 91,3                                                 & 91,5                                                           & 93,6         & 80           & 100          \\ \cline{1-4}
1             & DeBERTa                                              & \textbf{87,601}                                               & 90,3                                                         & 90,4           & 95,7                                            & 97,6                                             & 96,65                                                      & 98,4          & 88,2                                                  & 63,7                                                   & 75,95                                                            & 94,5                                                & 94,1                                                 & 94,3                                                           & 93,2         & 77,5         & 95,9         \\ \cline{1-4}
2             & T5 + Meena                                           & \textbf{87,097}                                               & 90,2                                                         & 91,3           & 95,8                                            & 97,6                                             & 96,7                                                       & 97,4          & 88,3                                                  & 63                                                     & 75,65                                                            & 94,2                                                & 93,5                                                 & 93,85                                                          & 92,7         & 77,9         & 95,9         \\ \cline{1-4}
4             & T5                                                   & \textbf{86,567}                                               & 89,3                                                         & 91,2           & 93,9                                            & 96,8                                             & 95,35                                                      & 94,8          & 88,1                                                  & 63,3                                                   & 75,7                                                             & 94,1                                                & 93,4                                                 & 93,75                                                          & 92,5         & 76,9         & 93,8         \\ \cline{1-4}
6             & PAI Albert                                           & \textbf{85,784}                                               & 86,1                                                         & 88,1           & 92,4                                            & 96,4                                             & 94,4                                                       & 91,8          & 84,6                                                  & 54,7                                                   & 69,65                                                            & 89                                                  & 88,3                                                 & 88,65                                                          & 88,8         & 74,1         & 93,2         \\ \cline{1-4}
5             & \begin{tabular}[c]{@{}l@{}}NEZHA\\ Plus\end{tabular} & \textbf{82,294}                                               & 86,7                                                         & 87,8           & 94,4                                            & 96                                               & 95,2                                                       & 93,6          & 84,6                                                  & 55,1                                                   & 69,85                                                            & 90,1                                                & 89,6                                                 & 89,85                                                          & 89,1         & 74,6         & 93,2         \\ \hline
\end{tabular}

\end{sidewaystable}

\end{document}